# Face Detection in Extreme Conditions: A Machine-learning Approach

Sameer Aqib Hashmi
*Department of Electrical and Computer Engineering*
*North South University, Bashundhara Dhaka, Bangladesh*
email: sameer.aqib@northsouth.edu

*Abstract-* Face detection in unrestricted conditions has been a trouble for years due to various expressions, brightness, and coloration fringing. Recent studies show that deep learning knowledge of strategies can acquire spectacular performance inside the identification of different gadgets and patterns. This face detection in unconstrained surroundings is difficult due to various poses, illuminations, and occlusions. Figuring out someone with a picture has been popularized through the mass media. However, it's miles less sturdy to fingerprint or retina scanning. The latest research shows that deep mastering techniques can gain mind-blowing performance on those two responsibilities. In this paper, I recommend a deep cascaded multi-venture framework that exploits the inherent correlation among them to boost up their performance. In particular, my framework adopts a cascaded shape with 3 layers of cautiously designed deep convolutional networks that expect face and landmark region in a coarse-to-fine way. Besides, within the gaining knowledge of the procedure, I propose a new online tough sample mining method that can enhance the performance robotically without manual pattern choice.*

## I. Introduction

Within the last numerous years, numerous algorithms have been proposed for face detection. While lots of development has been made in the direction of spotting faces below small versions in lighting fixtures, facial expression and pose, dependable techniques for popularity beneath greater excessive variations have been established elusive. Face detection is essential to many face applications, together with face popularity and facial expression evaluation. But, the huge visible versions of faces, together with occlusions, massive pose variations, and excessive lighting, impose splendid demanding situations for those obligations in actual-world applications.

Cutting-edge deep gaining knowledge of structures is proved to be nearly perfect face detectors, which outperform human abilities in this location [10]. The range of packages in these days' existence increases tremendously due to this reality. They would replace humans in regions wherein their accuracy is the most useful, as an instance, security. So given that their set of rules-driven selections may have critical results, the query of reliability and robustness towards malicious moves turns into critical. One in each of these undertakings is face detection that is widely used as a preparation operation for Face Id, which lets in tracing criminals or manage entrance policy.

The cascade face detector proposed by using viola and jones [2] makes use of haar-like features and AdaBoost to teach cascaded classifiers, which obtain correct performance with real-time performance. But, quite a few works [1, 3, 4] suggest that this detector may degrade substantially in real-global applications with large visual versions of human faces despite extra superior features and classifiers. Recently, convolutional neural networks (CNN) achieved incredible progress in a selection of computer vision tasks, inclusive of picture class [5] and face popularity [6]. Li et al. [8] use cascaded CNNs for face detection, however, it requires bounding box calibration from face detection with a more computational price and ignores the inherent correlation among facial landmarks localization and bounding field regression. Yang et al. [7] train deep convolutional neural networks for the facial characteristic reputation to reap excessive response in face areas which further yield candidate windows of faces. However, because of its complicated CNN shape, this technique is time-consuming in practice. However, most of the available face detection techniques forget about the inherent correlation among these responsibilities. Even though there exist several works that try to at the same time remedy them, there are nevertheless barriers in these works. However, the handcraft features used limits its overall performance. Zhang et al. [9] use multi-undertaking CNN to enhance the accuracy of multi-view face detection, but the detection accuracy is restrained using the preliminary detection home windows produced through a weak face detector. Alternatively, inside the education technique, mining tough samples in training are important to reinforce the energy of the detector. But, conventional difficult sample mining usually performs an offline manner, which considerably will increase the guide operations. Its miles are appropriate to design a web tough pattern mining approach for face detection and alignment, which is adaptive to the current schooling manner mechanically. With the rapid improvement of generation, face popularity is extra handy than different human frame popularity systems including fingerprints, irises, and DNA. It does no longer require compulsory participation and may resolve troubles without affecting human beings' regular existence. It has the advantages of low fee, high consumer popularity, and high reliability, and has vast application possibilities in identification, security monitoring, human-computer interaction, and different fields. The conventional face reputation method includes four ranges: face detection, face alignment, feature extraction, and face classification. The maximum crucial level is feature extraction, which immediately affects the accuracy of recognition. At present, in the restricted surroundings, the traditional neural network method has better outcomes in face popularity, however within the unrestricted environment, because of the complexity of the face photo leading to large intra-magnificence adjustments, as well because the inter-class adjustments due to the outside light and heritage, the traditional neural network face popularity method regularly fails to obtain the desired effects.

A mature face detecting system commonly consists of image acquisition, photo pre-processing, face detection, face tracking, face alignment, function extraction, and evaluation. Among the extra critical steps are face detection, tracking, and face characteristic extraction. In recent years, face reputation structures were extensively used in channel bayonet structures which include clever access management and identity verification in high-pace railway stations. These channel bayonet face recognition structures have all or maximum of the face picture series, face detection, face alignment, face high-quality detection, face function extraction, face tracking, and different steps. However, some of those structures require an excessive degree of cooperation from people, a few are complex to put in force, and a few have high necessities for hardware along with computing gadgets. On the only hand, the computing power of embedded systems isn't enough to support face detection, tracking, and face feature pairing-based totally on deep getting to know. Real-time requirements, a few channel bayonet face recognition systems require humans to intentionally approach the camera to cooperate with the device for verification, discarding the herbal and convenient benefits of face recognition. The specific goal situation studied in this paper is a single channel bayonet (unmarried face near range), and the aim is so one can quickly examine and recognize faces within 1-4 meters. The purpose of the research is to use a faster and higher overall performance algorithm to the channel bayonet face reputation machine with low computing energy and to enhance the running speed of the face popularity gadget through the progressed face detection algorithm. It can be established on low-stop gadgets with terrible computing overall performance even as preserving sure detection and reputation overall performance.

In this paper, I propose a new framework to integrate these tasks into the usage of unified cascaded CNNs via multi-assignment learning. The proposed CNNs consist of three parts. Within the first stage, it produces candidate windows quickly via a shallow CNN. Then, it refines the windows to reject a huge range of non-faces home windows through an extra complicated CNN. Ultimately, it uses a greater effective CNN to refine the end result and output facial landmarks positions. Way to this multi-mission getting to know the framework, the performance of the set of rules may be substantially progressed. The predominant contributions of this paper are summarized as follows: (1) I recommend totally new cascaded CNN's primarily based framework for face detection, and thoroughly layout light-weight CNN structure for real-time performance. (2) I suggest a powerful technique to behave on-line difficult pattern mining to enhance the overall performance. (3) Great experiments are carried out on difficult benchmarks, to expose the giant overall performance improvement of the proposed method in comparison to the latest strategies in each Face Detection responsibilities.

II. BASIC PRINCIPLES

*A. Face detection*

A multi-task cascaded convolutional network (MTCNN) is a framework developed as an answer for both face detection and face alignment. The manner includes 3 degrees of convolutional networks that can apprehend faces and landmark places which include eyes, nostrils, and mouth. The paper proposes MTCNN as a way to integrate both tasks (reputation and alignment) and the usage of multi-challenge studying. Inside the first degree, it uses a shallow CNN to quickly produce candidate home windows. Inside the 2d level, it refines the proposed candidate home windows through a greater complicated CNN. And lastly, inside the third stage, it makes use of a third CNN, extra complex than the others, to similarly refine the result and output facial landmark positions.

MTCNN is a method of face detection and alignment primarily based on deep convolutional neural networks [11], [12] that is to mention, this technique can accomplish the assignment of face detection and alignment at the same time. In comparison with the traditional approach, MTCNN has better overall performance, can appropriately discover the face, and the speed is likewise faster, besides, MTCNN also can hit upon in real-time. MTCNN consists of 3 neural network cascades, particularly p-net, r-net, and o-internet. With the intention to attain face reputation on a unified scale, the authentic picture must be scaled to one of a kind scale to shape a photo pyramid before the usage of these networks. The first community p-internet is a complete convolutional community used to generate candidate window and border regression vectors. Bounding field regression is used to correct candidate bins, and then non-maxima are used to suppress these combined overlapping candidate boxes.

Step one is to take the image and resize it to one-of-a-kind scales for you to construct an image pyramid, which is the input of the subsequent 3-stage cascaded community. The structure of the P-Net is shown below.

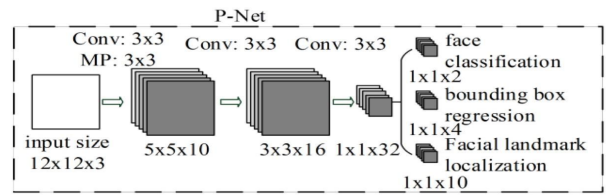

Fig. 1. P-Net network structure

All candidates from the P-Net are fed into the refining community. The word that this network is a CNN, not an FCN like the one before due to the fact there's a dense layer at the closing stage of the network structure. The R-Net further reduces the number of applicants, plays calibration with bounding field regression, and employs non-maximum suppression (NMS) to merge overlapping applicants. The r-internet outputs whether the input is a face or not, a 4 detail vector that is the bounding field for the face, and a 10 element vector for facial landmark localization.

This stage is just like the R-Net, however, this output network aims to describe the face in more detail and output the five facial landmarks' positions for eyes, nostril, and mouth.

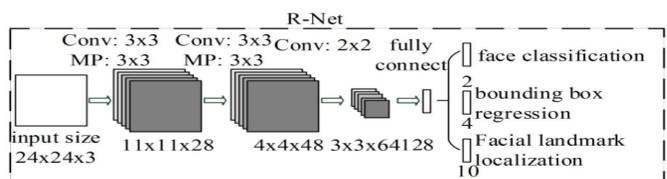

Fig. 2. R-Net Network Structure

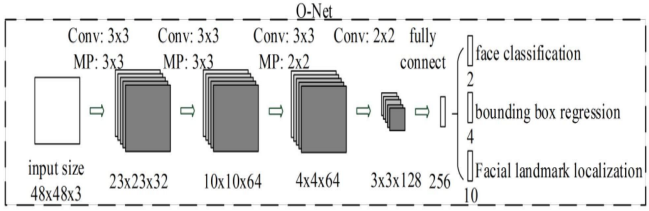

Fig. 3. O-Net network structure

Each network in MTCNN has three parts of the output, so the loss is also composed of three parts. For the face detection part, directly use the cross-entropy loss function:

$$L_i^{det} = - (y_i^{det}\log(p_i)+(1-y_i^{det})(1-\log(p_i))) \quad (1)$$

In the formula, $p_i$ represents the probability of inputting a face, and $y_i^{det}$ det represents a true label. The box regression and the five feature point decisions are all regression problems, so use the common Euclidean distance to find the loss. Boundary box regression loss function:

$$L_i^{det} = \| \hat{y}_i^{box} - y_i^{box} \|_2^2 \quad (2)$$

Where $\hat{y}_i^{box}$ is predicted by the network and $y_i^{box}$ is the actual real background coordinates. Key point decision loss function:

$$L_i^{Landmark} = \| \hat{y}_i^{Landmark} - y_i^{Landmark} \|_2^2 \quad (3)$$

Where $\hat{y}_i^{Landmark}$ is predicted by the network and $y_i^{Landmark}$ is an actual real key point coordinate. Finally, the three losses are multiplied by their own weights and then added together to form the final total loss.

## B. Convolutional Neural Network

Convolutional neural networks are a sort of feedforward neural networks with convolutional computation and deep structure. They're one of the representative algorithms of deep getting to know. Convolutional neural network (CNN) extracts high-stage semantic statistics from raw statistics entered layer by layer through stacking a sequence of operations such as convolution, convergence, and nonlinear activation function11. In convolutional neural networks, the function of the convolutional layer is to educate fewer parameters to extract function facts for the entered records. The biggest gain of the convolutional layer in comparison with the overall connection is that the community is regionally related, and the quantity of parameters that want to be trained is small, which is conducive to building a deeper and larger network shape to solve greater complicated troubles. The position of the pooling layer is to lessen the dimensions of the characteristic map. To speed up the network training and reduce the number of computational facts, the convolutional neural community makes use of a pooling layer in the back of the convolutional layer to reduce the quantity of information, the pooling operation can't most effective make the characteristic size extracted by the convolution layer smaller, lessen the number of computing records, but additionally reduce the diploma of over-becoming of the network to some extent and enhance community overall performance. The function of the fully related layer is to map the feature map of a two-dimensional photo onto a one-dimensional feature vector. Through the overall connection, the function map of any size can be mapped into the vector of the desired size [13], [14].

## III. LITERATURE REVIEW

One of the toughest and traumatic tasks is to enhance the accuracy of object detection within the computer vision and prescient discipline, inclusive of the human face and eyes. Researchers across the globe are working on this place that allows you to use the satisfactory-found objects in numerous packages. According to Kasinski [5], Haar cascade classifiers are becoming not unusual in face-quit eye detection. It characterizes an HCC-primarily based three-stage hierarchical face and eye detection device. HCC consists of 2500 advantageous facial expressions for identification of the face. There are 2900 images taken wherein there may be no call. Face detectors are equipped with a photo of 2500 left or right eyes and the snapshots of the eyestrain terrible sets. Overall advantageous 94 percent and fake-fantastic thirteen percent are detected in facial detection. Eyes are detected at a fee of 88 percentages with the simplest 1 percent false nice outcome.

Primarily based on deep convolutional network techniques, Zhang [11] adopted 3 ranges of deep convolutional networks that may predict the coarse-to-exceptional position of face and landmarks superbly. A current look at has proven that during this discipline, deep getting to know methods can have vast effects. The writer has advised CNNs for eye detection consisting of trio tiers: idea network (p-net), refinement community (r-net), and output network (o-net). Experimental consequences unearth these strategies to exceed trendy techniques over multiple disturbing assessments whilst keeping efficiency in actual-time.

Lang ye [15] counseled a singular CNN framework to boost the precision of eye detection at once making use of the uncooked color values of photo pixels with the aid of CNN. The primary factor senses rough bounding packing containers of capability eye patches. The second step decides whether or now not the tough bounding boxes belong to the eyes and exclude the non-eye bounding boxes. 8300 eye samples of various mild situations, resolutions had been received. Sooner or later, entire samples have been split into training and validation datasets of 500 samples in step with magnificence inside the validation set. The second level of CNN outperforms the first tier of CNN, accomplishing an accuracy rate of seventy-three percent and a keep in mind price of 76 percentages respectively.

## IV. Experimental Result

My commentary is done in a couple of different model classes. The primary experiment accomplished the usage of the MTCNN version, where I used my dataset. The second experiment is likewise accomplished using the Haar cascade set of rules on my dataset. To explain my proposed model, a comparative evaluation of my proposed face reputation models the usage of MTCNN and Haar cascade is proven after my dataset is carried out as a check facts series. Using MTCNN I got ninety-nine percent accuracy rates, and the usage of Haar cascade sixty-eight percent accuracy is performed. Then I compared my effects from each of my experiments according to precision, accuracy, and do not forget. Subsequently, I equate my paintings within the applicable discipline of object detection with the present works. Section a to c gives a comprehensive overview of my experimental consequences and a precis of my findings.

*A. Results of MTCNN & Haar Cascade Algorithm*

After making use of my dataset to the MTCNN procedure, I determined the face of the images for approximately a hundred videos at a rate of 99%-100%. Right here, the end result suggests that a great final result has been finished: the use of multi-venture cascaded Convolutional networks. "Fig. 4", presents a number of the pix from my dataset in which the face has been efficiently decided by the usage of the MTCNN manner.

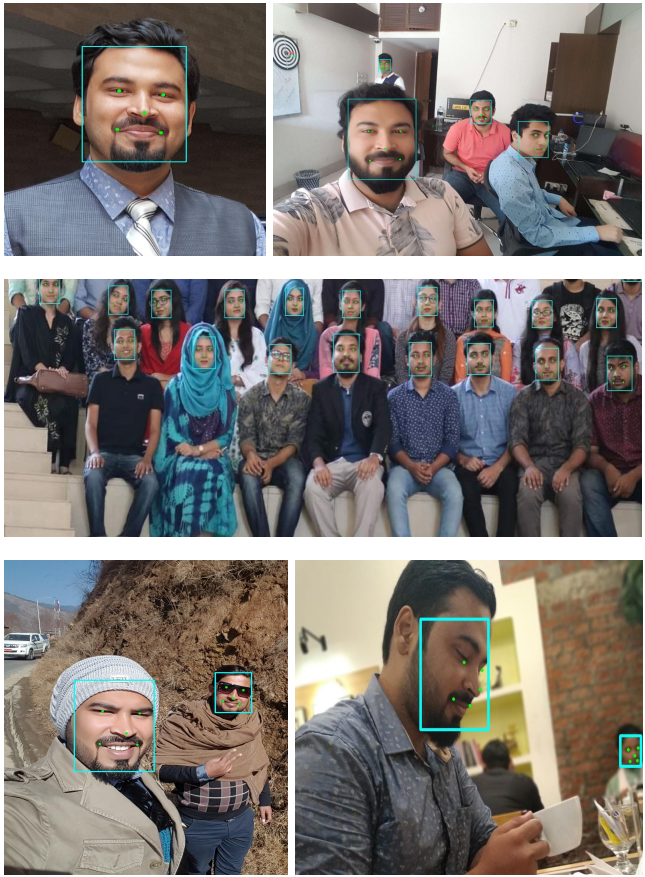

Fig. 4. Face and Facial Points Detection using MTCNN

After making use of my dataset to the Haar cascade approach, I determined the face at 68%. "Fig. 5", shows some of the pictures from my dataset where the face had been efficiently determined using the Haar cascade method.

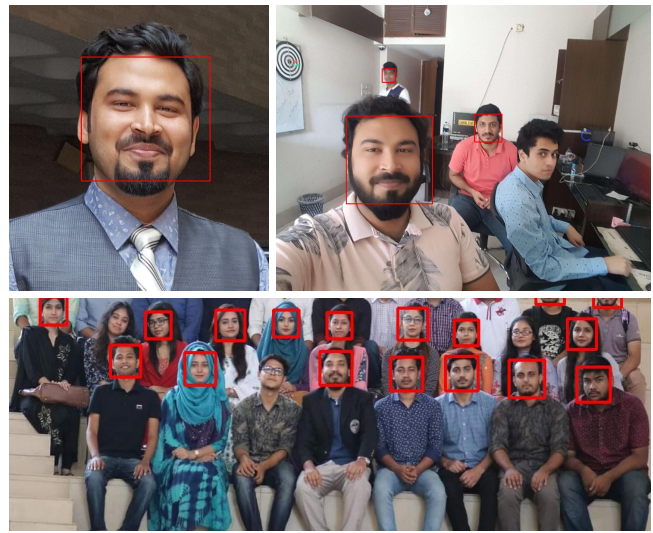

Fig. 5. Face Detection using Haar Cascade

*B. Result Comparison*

I can analyze the detection rates among these methods using various datasets in distinct methods. I have evaluated the difference in detection rate between the strategies in Table. I.

Table. I
COMPARISON OF DETECTION RATE OF FACIAL POINTS AMONG VARIOUS METHODS FOR DIFFERENT DATASETS

| Algorithm | Accuracy |
| --- | --- |
| Viola-Jones | 74.38% |
| Haar Cascade | 94% |
| MTCNN | 99.95% |

In comparison, the huge amount of data set Haar Cascade gave better accuracy. The MIT dataset has been used for Viola-Jones, 10000 sets of images for Haar Cascade and For MTCNN the default trained library has been used. That's why this accuracy was so perfect that sometimes it's detection rate becomes 100%.

After using the dataset in extraordinary techniques I am able to examine the detection charges amongst these strategies. Inside the Table. II, I have analyzed the difference in the detection rate of many of the methods.

Table. II
COMPARISON OF DETECTION RATE OF FACIAL POINTS AMONG VARIOUS METHODS FOR MY OWN DATASETS

| Algorithm | Accuracy |
| --- | --- |
| Haar Cascade | 68.16% |
| MTCNN | 98% |
| Viola-Jones | 61.81% |

Various datasets are used to compare different methods. Hence, drawing a comparison chart Fig. 6 and Fig. 7 can compare the detection rate between my own dataset and different datasets for different methods.

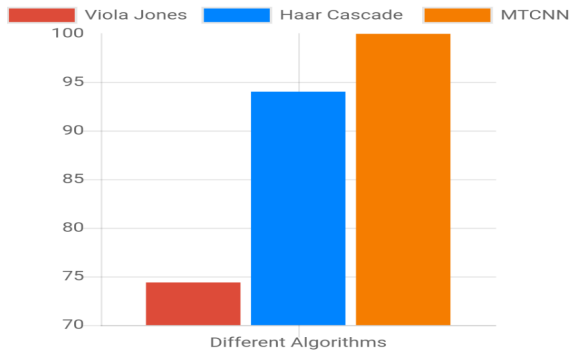

Fig. 6. Comparison Among Different Algorithms for Different Datasets

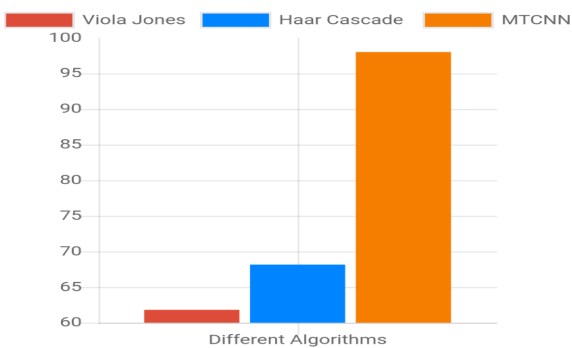

Fig. 7. Comparison Among Different Algorithms for My Different Datasets

*C. Analysis & Discussion*

I have calculated the confusion matrix of eyes and face detection for my proposed method MTCNN using my dataset.

Table. III

CONFUSION MATRIX OF FACIAl POINTS DETECTION(MTCNN)

|  | Positive | Negative |
|---|---|---|
| **Predicted Positive** | 17620 | 335 |
| **Predicted Negative** | 30 | 280 |

It is drawn in Table. III. The confusion matrix has been generated from 100 videos containing about 20000 images.

Table. IV

CALCULATION OF CONFUSION MATRIX FOR FACIAL POINTS DETECTION

| Measure | Value(%) | Derivation |
|---|---|---|
| Precision | 98.13 | PPV=TP/(TP+FP) |
| Recall | 99.83 | TPR=TP/(TP +FN) |
| Specificity | 45.53 | FPR=FP/(FP+TN) |
| F-Measure | 98.97 | F1 = 2TP / (2TP + FP + FN) |
| Accuracy | 98.00 | TP+TN / TP+TN+FP+FN |

And in line with Table. IV the accuracy charge is 98%. In which the precision is 98. 13% and keep in mind is 99.83%. And ultimately, the F-1 score is 98.97%. After evaluating all of the comparisons it's clear that my records are higher for deep gaining knowledge of techniques (MTCNN) to Detect faces. Consequently, given all of the one-of-a-kind strategies for my very own dataset and different datasets, the overall performance inside the deep learning method was pleasant.

IV. CONCLUSION

In this paper, I eventually proposed a framework for face detection based on multi-task cascaded CNNs. Experimental results and my other proposed framework show that my MTCNN techniques consistently outperform most of the main techniques, thinking about my dataset as a check dataset. There are plenty of programs that could use my proposed technique like my dataset when you consider that I can remember the variety of people in a scene by means of image popularity of the face. In the future, I am able to try to work on different eyes and face studies activities together with facial features or mind-set identity, detection of weakness, motion of the iris and detection of a questionable observer. I have got a plan to paint on these responsibilities through the usage of my dataset so I will use the statistics set nicely to teach my version for performance improvement.